\documentclass[letterpaper, 10 pt, conference]{ieeeconf}  % Comment this line out if you need a4paper

\IEEEoverridecommandlockouts                              % This command is only needed if 
\usepackage{graphics} % for pdf, bitmapped graphics files
\usepackage{times} % assumes new font selection scheme installed
\usepackage{amsmath} % assumes amsmath package installed
\usepackage{amssymb}  % assumes amsmath package installed
\usepackage{graphicx}
\usepackage{booktabs}       % professional-quality tables
\usepackage{color}

\title{\LARGE \bf
Just-in-Time Reconstruction: \\ Inpainting Sparse Maps using Single View Depth Predictors as Priors
}
\author{Chamara Saroj Weerasekera$^{1}$,  Thanuja Dharmasiri$^{2}$, Ravi Garg$^{1}$, Tom Drummond$^{2}$ and  Ian Reid$^{1}$% <-this % stops a space
\thanks{ $^1$ School of Computer Science, The University of Adelaide, Australia}
\thanks{ $^2$ Department of Electrical and Computer Systems Engineering, Monash University, Australia}
\thanks{ $^{1,2}$ Australian Center for Robotic Vision}% <-this % stops a space
}

\begin{document}

\maketitle
\thispagestyle{empty}
\pagestyle{empty}
%%%%%%%%%%%%%%%%%%%%%%%%%%%%%%%%%%%%%%%%%%%%%%%%%%%%%%%%%%%%%%%%%%%%%%%%%%%%%%%%
\begin{abstract}
%\textcolor{blue}{FIXME:Ravi toggle red/blue}

We present ``just-in-time reconstruction" as real-time image-guided inpainting of a map with arbitrary scale and sparsity to generate a fully dense depth map for the image. In particular, our goal is to inpaint a sparse map --- obtained from either a monocular visual SLAM system or a sparse sensor --- using a single-view depth prediction network as a virtual depth sensor. 
%
%commented from the old abstract
%In this work we present a novel approach to fuse noisy depth maps that could be of arbitrary scale and sparsity. In particular, we apply our method to perform image-guided sparse depth map interpolation where a dense depth map predicted from a RGB image using a neural network is fused with a sparse depth map coming from either multi-view geometry or a sensor.}
%
We adopt a fairly standard approach to data fusion, to produce a fused depth map by performing inference over a novel fully-connected Conditional Random Field (CRF) which is parameterized by the input depth maps and their pixel-wise confidence weights. Crucially, we obtain the confidence weights that parameterize the CRF model in a data-dependent manner via Convolutional Neural Networks (CNNs) which are trained to model the conditional depth error distributions given each source of input depth map and the associated RGB image. Our CRF model penalises absolute depth error in its nodes and pairwise scale-invariant depth error in its edges, and the confidence-based fusion minimizes the impact of outlier input depth values on the fused result.
% added by saroj:
We demonstrate the flexibility of our method by real-time inpainting of ORB-SLAM, Kinect, and LIDAR depth maps acquired both indoors and outdoors at arbitrary scale and varied amount of irregular sparsity.
\end{abstract}

%The proposed fully-connected CRF energy is quadratic and continuous and can therefore be computed and inferred efficiently in linear-time. We show results for ***FIXME*** demonstrating the efficiency of our approach, including real-time depth infilling for Kinect and ***FIXME***. 

\section{Introduction}
Simultaneous Localization and Mapping (SLAM) is a well studied problem and is a backbone of many visual robotics systems. State-of-the-art large scale SLAM successfully integrates information from multiple views to build a consistent map while also accounting for the map's uncertainty. 
However, efficiently estimating and storing large scale, dense maps is a challenging problem. A compromise on the map's detail is often a necessity in favour of scalable optimization and efficient storage of the map. While dense SLAM methods like DTAM \cite{Newcombe2011} are capable of real-time dense reconstruction of key-frames, methods using a point cloud representation for sparse or semi-dense mapping \cite{ORBSLAMRSS_2015,ORBSLAM_2015,Engel2014,Klein2007} still remain a popular choice for large scale mapping and tracking.
%Most popular large scale SLAM systems rely on semi-dense representation of the maps with 3D point cloud.

%\textcolor{blue}{a couple of lines goes hear saying why SLAM is something we care for?}

Although these semi-dense maps are very useful for tracking and localizing a robot in an office or even a autonomous vehicle in a city, a richer representation is often desired for successful navigation and for robot to interact with the environment once it has located itself in the map. In this work we advocate a more flexible approach where a denser representation of a small portion of the map can be rapidly generated from a sparse 3D point cloud \emph{on-demand} by a task requiring richer understanding of the scene -- we refer to this as ``just-in-time reconstruction".

%\textcolor{red}{FIxME: Not sure how to justify use of single view depth prediction networks and disadvantage of Crf in here} 
Image-guided inpainting using hand-crafted locally connected \cite{Colorization,CrossBilateralFiltering} or fully connected Gaussian \cite{Krahenbuhl2012} CRFs with learnable pairwise potentials (on depth and image colors) presents itself to be an obvious choice for the task. However, these CRFs ignore the arbitrary scale, varied irregular sparsity and/or uncertainty at which visual SLAM maps are created. Moreover they are very restrictive, relying on hand-crafted priors that link image colours and textures to depths. 

On the other hand, in recent years, CNNs have emerged as immensely successful tool to directly learn the highly non-linear relationships between RGB images and depth in a more flexible way \cite{Eigen2014,Depth2015Liu,laina2016deeper,Ummenhofer2016,dharmasiri2017joint, KendallEndToEnd17} through supervised, or even unsupervised \cite{garg2016unsupervised} means. %From saroj to ravi: i changed some wordings in this para you wrote, pls check.
%While supervised CNNs like \cite{Eigen2014,Depth2015Liu,laina2016deeper,Ummenhofer2016,KendallEndToEnd17} now are capable of generating vivid reconstructions from a single or couple of indoor image, \cite{garg2016unsupervised} are capable of using standard stereo/monocular geometry to achieve very plausible reconstructions in outdoor environments even without direct supervision.
%These methods however being real-time only rely on single frame or a couple \cite{Ummenhofer2016,KendallEndToEnd17} and little work has been done to integrate them into a SLAM framework.
While CNNs are not as easily scalable as most visual SLAM methods to incorporate information from hundreds of images at inference time, our view is that these depth prediction networks, in the form of millions of learnable weights, naturally allow for a much richer understanding about the distributions of valid partial maps conditioned on image patches over a large set of correlated images. Here we explore how to take advantage of this rich and flexible representation for ``just-in-time reconstruction" to densify a sparse map that already incorporates complementary information from multi-view geometry or a depth sensor.
% \textcolor{red}{Maybe fusion here? followed by the para saroj had in old intero listing the problem.}
%However, the main caveats for single view reconstruction... most approaches at test time use one or few images, not straight forward to extend to multiple views, difficult to fuse with other sensory data, is No association with the stored maps which we use for localization. \textcolor{blue}{**FIXME : CNNs to be used as prior}

\begin{figure*}
\centering
\includegraphics[width=\textwidth, trim={0cm 0cm 0cm 15px},clip]{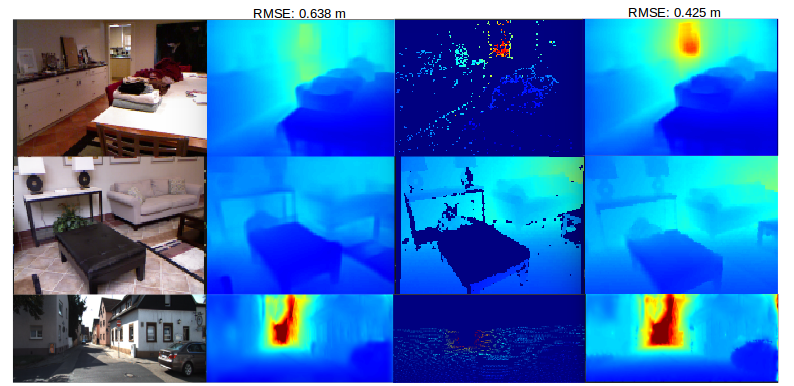}
\caption{Three different scenarios where our method can be applied. Top Row to Bottom Row: Image-guided inpainting of an ORB-SLAM depth map (obtained at an arbitrary scale consists outliers and is highly sparse), a Kinect depth map (with structured holes), and a LIDAR depth map (of limited range and coverage) respectively. From left to right, in each row consists: the input image, single view depth prediction results, sparse/incomplete depth map obtained using SLAM method or a sensor, our fused dense depth maps.} %Note that in the first scenario ORB-SLAM's depth map is correct only upto a scale factor and consists of outliers, both issues which are taken of by the probabilistic scale-invariant fusion, without any depth pre-scaling or hard-coded outliers removal.}
\label{fig:intro}
\end{figure*}

More formally, we present ``just-in-time reconstruction" as a novel %saroj: added approach to
approach to real-time image-guided inpainting of a point-cloud obtained at an arbitrary scale and sparsity, to generate a fully dense depth map corresponding to the image. In particular, our goal is to inpaint a sparse map -- obtained from either a monocular visual SLAM system or a sparse sensor -- using a neural network which predicts dense depths from a single RGB image as a virtual sensor. In other words we want to probabilistically fuse a sparse SLAM map with information from a single-view depth prediction neural network. %\textcolor{blue}{added by saroj: is this true? - since our crf is parameterized by learned CNN depths it brings the inpainting problem closer to data fusion. Can we simply establish the link of inpainting and fusion by saying this line?}\textcolor{red}{Ravi: I know what you are trying to say but it requires some refinement}

%\textcolor{blue}{Not sure about this para here} 
This fusion process is however non-trivial due to structured errors present in single view depth predictions as well as the irregular sparsity and arbitrary scale of the SLAM map. For instance, depth maps from monocular visual SLAM are typically accurate at regions corresponding to high image gradient but correct only \emph{up-to-scale}, and are usually irregularly sparse and can contain outliers. On the other hand depth predictions from a CNN are usually in metric scale, and dense, but have varying amount of errors depending on the capacity and generalization capability of the network used. Similarly depth maps from dedicated depth sensors, while being accurate, still suffer from the limited range and coverage of the sensors. %\textcolor{blue}{some jump here if this para is used}

Adapting standard practices of probabilistic data fusion, we formulate just-in-time reconstruction as an inference problem over a novel learnable fully-connected Conditional Random Field (CRF). Nodes of this CRF penalize absolute depth error of the reconstructed depth map from the sparse observations stored in a large map while the edges enforce the pair-wise scale invariant depth relations (in this work, depth ratios of any two points in the image) of the inferred depth map to be the same as that of a single-view depth predictor while accounting for per-point confidence in the sparse map as well as confidence in the dense depth predictions. 

Particularly, we obtain the confidence weights that parameterize the CRF model in a data-dependent manner via Convolutional Neural Networks (CNNs) which are trained to model the conditional depth error distributions given each source of input depth map and the associated RGB image.
Section \ref{method} outlines our novel CRF in detail which (i) while being flexible enough to account for prediction uncertainties of sparse as well as dense maps (ii) allows for linear-time inference in real-time. 
%for learned uncertainty Section \ref{},\ref{} 

Our CRF formulation is very general and can be used to probabilistically fuse an arbitrary number of (sparse or dense) depth maps obtained from multiple input sensors. We also extend uncertainty estimation methods like in \cite{kendall2017uncertainties,Ummenhofer2016} to predict confidences for sparse and dense maps obtained at arbitrary scale (section \ref{learning}) for use as parameters of our CRF.

We demonstrate effectiveness of the proposed ``just-in time-reconstruction" approach by inpainting sparse maps with varied irregular sparsity and scale variations. In particular we show inpainting of (i) ORB-SLAM depth maps with scale ambiguity, outliers and highly irregular sparsity (ii) partially incomplete depth maps of indoor scenes obtained by a Kinect sensor and (iii) sparse point clouds obtained outdoors using a LIDAR sensor which have limited range and density, guided by single-view depth predictions and learned pointwise confidences. Figure \ref{fig:intro} presents a snapshot of these results. Our just-in-time reconstruction approach generates depth maps with rich structural details than the semi-dense point clouds and gives more accurate reconstructions outperforming state-of-the-art image guided inpainting baselines and single-view depth prediction networks.

\section{Just-in-time reconstruction}
\label{method}
%\textcolor{blue}{We need following subsections}
%1) Problem Definition and Notation.
%2) Loss function 
%3) node of crf - push the inferred depth to be same as sparse map
%4) Edges (fully connected bit) -> Fusion of the relative depth ratios
%5) Local connections to further inpaint

%\subsection{Problem Definition and Notation}
We define just-in-time reconstruction as an inference over a fully connected learnable CRF to inpaint a sparse map with $P$ points that has been aligned with a single RGB image $I$ with $N$ pixels and represented as a \emph{partial} log-depth map $y^s = [y_1^s, .., y_N^s] = ln(d^s) = [ln(d_1^s), .., ln(d_N^s)]$, where $d_i^s$ is the depth of pixel $i$, with valid depths only at the projected pixel locations.
Assuming that $c^s=[c_1^s, .., c_N^s], 0 \leq c_i^s \leq 1, \forall {i}$ is the confidence associated with the map (0 for invalid depths) estimated by a probabilistic SLAM approach or learned using a neural network trained to predict map confidence from input data (explained in section \ref{learning}), our goal is to infer the dense log-depth map $y = [y_1, .., y_N] = ln(d) = [ln(d_1), .., ln(d_N)]$. 
We assume that for this inpainting task we are given with a dense single view depth prediction network and a data-driven depth confidence prediction network (section \ref{learning}). We denote the log-depths regressed by the depth prediction network to be $y^{d} = [y_1^{d}, .., y_N^{d}]$,
and respective confidence maps to be $c^{d}=[c_1^{d}, .., c_N^{d}], 0 \leq c_{i}^d \leq 1$.

To achieve the ``just-in-time reconstruction" from an image as described above we propose to minimize the following CRF energy w.r.t. $y$:
\begin{equation} 
\label{energy_inference}
\begin{aligned}
 E(y) = \alpha E_u(y,y^s,c^s) + \beta E_{fc}(y,y^d,c^d) + \gamma E_{lc}(y,y^d,c^d) \\
\end{aligned}
\end{equation}
where, $E_{u}$ is the unary term generating the log depths for image $I$ consistent with the sparse map, $E_{fc}$ is a fully-connected pairwise term and $E_{lc}$ is a locally connected pairwise term, both penalizing incorrect pairwise depth relationships (depth ratios as scale invariant measures) of the inferred dense log depth map using the single view depth predictions as the learned priors.\footnote{In a general form multiple sparse and dense depth maps obtained at arbitrary scale from various sources can be used in our framework, replacing the CNN-based single view depth prediction, and simply minimizing the sum of the pairwise and fully connected terms described above, for each given depth map, in order to inpaint the desired sparse map.} The tunable parameters $ (\alpha, \beta, \gamma) >0 $ signify the relative importance of each term.
A detailed description of each CRF term, the motivation behind using them and relations of these to existing frameworks are described in the following subsections.

\subsubsection{Nodes of CRF}
The unary term in the CRF pulls the inferred depth map to be consistent with the sparse map obtained via SLAM or a sensor. Inspired by \cite{LadickyPullingThingsOutofPerspective_CVPR2015,EigenPF14,Eigen2014}, we use squared natural log-depth differences for every point on the sparse map as the unary potentials of our just-in-time reconstruction CRF:
\begin{equation} 
\label{energy_unary}
\begin{aligned}
E_u(y) = \sum_i^N  c_i^s (y_i - y_i^s)^2 \\
\end{aligned}
\end{equation}
Notice that each term in $E_u$ is weighted by the learned confidence of map accuracy $c_i^s$ ($c_i^s=0$ if no depth). The log-depth parameterization and $c_i^s$s increase the model's homoscedasticity, favoring the least squares estimator.

\subsubsection{Edges of fully connected  CRF model}
As our goal is to inpaint sparse maps of arbitrary scale we aim to design a learnable prior which is insensitive to the scale of the scene. For this purpose, we propose the pairwise potentials of our full connected CRF to be:
\begin{equation} 
\label{energy_fc}
\begin{aligned}
E_{fc}(y) = \frac{1}{2N}\sum_{i,j} c_{ij}^d \big((y_j - y_i) - (y_j^d - y_i^d)\big)^2 \\
\end{aligned}
\end{equation}
where $c_{ij}$ are the learnable parameters of our CRF intuitively representing correctness of the log of depth-ratio $ln(d_j/d_i) = y_j^d - y_i^d$ of the single view depth predictions for any two points $i$ and $j$ to be learned in a data driven fashion.

Pairwise terms of our fully connected CRF can be best interpreted as the terms which enforce scale-invariant ordinal relationships of the inferred depths of two points to be same as that of the single view depth prediction network.
The intuition behind using this fully connected CRF is that depth ratios of two points in a scene are invariant to the scale of the scene. Any other scale invariant function $f(d_i,d_j)$ may be used without loss of generality in our framework in place of $ln(d_j/d_i)$.

The fully connected pairwise CRF defined in \eqref{energy_fc} is however intractable in its most generic form, as the number of learnable parameters $c_{ij}^d, \forall(i,j)$ grows quadratically with the number of pixels in the image. Approximations are generally used to model $c_{ij}^d$ in parametric form for reducing the number of independent learnable parameters and for efficient inference. The most common practice is to model $c_{ij}^d$ as the sum of Gaussian RBF kernels each having two learnable parameters, which are mean and variance. For example \cite{FastBilateralSolver} and \cite{Krahenbuhl2012} define $c_{ij}^d$ in the form of Gaussian RBF kernels that are a function of the distance between pixel $i$ and $j$ and the color difference between those pixels. These CRF models allow for fast inference but are very restrictive.

In this work, we propose a different relaxation to $c_{ij}^d=c_i^dc_j^d$ which allows for efficient inference while having many more learnable parameters for expressiveness. The intuition is that the accuracy of the pairwise term is limited by the least confident depth value forming the ratio, and thus the overall confidence can be approximately expressed as a product of individual ones. This simple approximation significantly reduces the number of parameters to learn, and also allows for tractable inference (refer section \ref{inference}) as the fully connected term in equation \eqref{energy_fc} (and thereby its gradient) can now be re-written in an alternative form that allows for linear time computation:

\begin{equation} 
\label{energy_fc_lt}
\begin{aligned}
E_{fc}(y) &= \frac{1}{N} \textstyle{\sum}_j^N c_j^d \displaystyle \sum_i^N  c_i^d (y_i - y_i^d)^2 \\ &- \frac{1}{N} \Big(\sum_i^N c_i^d (y_i - y_i^d)\Big)^2 \\
\end{aligned}
\end{equation}

\subsubsection{Local Grid Connected Edges of CRF}
Additionally, we define a grid connected term of the CRF to give more importance to the local structures learned by the single view depth predictor:
\begin{equation} 
\label{energy_lc}
\begin{aligned}
E_{lc}(y) = \sum_{i,k} c_i^dc_k^d \big((y_k - y_i) - (y_k^d - y_i^d)\big)^2 \\
\end{aligned}
\end{equation}
where $k\in{\{+u(i),+v(i)\}}$ denotes pixel locations to the right of and below pixel $i$ in the image plane. This enforces the solution to trust pairwise relations in $y^d$ mainly around a local neighborhood around each pixel $i$, and is thus helpful for providing local support for the unary term where depth information is absent, where the solution is otherwise increasingly biased towards $y^d$ as information from the dense fully connected terms dominates the weak information of $y^s$ carried over from the unary potentials.

\begin{figure*}[t]
\centering
\includegraphics[width=\textwidth]{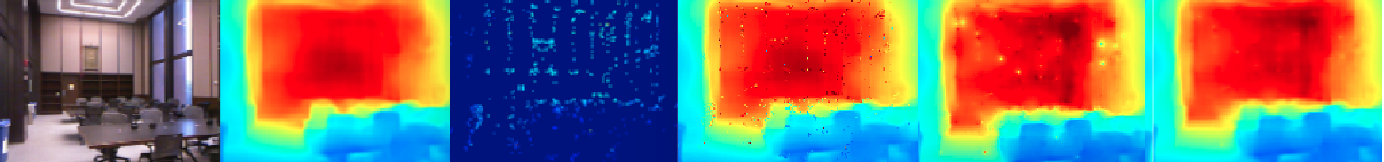}
\caption{Qualitative ablation of the impact of data-driven smoothing as defined by our local CRF pairwise terms $E_{lc}$, and the impact of incorporating confidences into the energy. In column order  $1^{st}$: RGB Image, $2^{nd}$: Dense depth prediction of Eigen \cite{Eigen2014} (metric scale), $3^{rd}$: Sparse depth map of \cite{ORBSLAM_2015} (arbitary scale), $4^{th}$: Our reconstruction with learned confidences but without $E_{lc}$, $5^{th}$: Our reconstruction with $E_{lc}$ but without learned confidences,  $6^{th}$: Our final reconstruction with both $E_{lc}$ and learned confidences.}
\label{fig:fusions}
\end{figure*}

The end-effect of $E_{lc}$ is similar to a data-driven local smoothing (refer Figure. \ref{fig:fusions}) where information from $y^d$'s local pairwise depth relationships are used to ``smooth over'' the areas where the sparse map's points are fused in the solution, while still anchoring the solution onto the sparse map. Apart from computational efficiency, we only consider a 4-connected graph for $E_{lc}$ as $E_{fc}$ already encompasses the full pairwise connectivity graph and, in addition to incorporating pixelwise confidences, our formulation provides the flexibility of using the tunable weights $\beta$ and $\gamma$ to control how much the solution gets biased towards $y^s$ and $y^d$ for a fixed $\alpha$ (refer Figure \ref{fig:hyperparams}). We denote the set of pixels in the neighborhood of $i$ as $\mathcal{N}(i)$. Note that additional model expressibility can be added to our locally connected pairwise terms by increasing the size of $\mathcal{N}(i)$, and for instance introducing a multiplicative pairwise pixel-distance based Gaussian RBF kernel with tunable variance to the terms in equation \eqref{energy_lc} --- such that nearby pairwise depth ratio inconsistencies are penalized more strongly than those further apart --- while still retaining the same inference method.

% While the neighborhood size can be arbitrary, a neighbourhood that is too large can make the solution increasingly biased towards a particular $\mathbf{y}^d$, and in the limit $E_{lc}$ will become same as $E_{fc}$.

% Works such as \cite{Campbell2013}\cite{SURGE} use a Gaussian-RBF kernel parameterized by pixel distance to offer flexibility in reducing the pairwise weight with increasing pairwise connectivity distance, but requires additional pre-processing overhead. Our formulation although free of a pixel distance-based Gaussian-RBF kernels offer similar flexibility simply 

\subsection{Inference Method}
\label{inference}
%Combining \eqref{energy_unary}, \eqref{energy_fc_lt}, and \eqref{energy_lc} into \eqref{energy_inference} we obtain:

%\begin{equation} 
%\label{energy_inference_long}
%\begin{aligned}
%E^{(m)}(\mathbf{y}) = & \frac{\alpha^{(m)}}{N_v^{(m)}}  \textstyle{\sum}_j^N \displaystyle c_j^{(m)} \sum_i^N  c_i^{(m)}(y_i - y_i^{(m)})^2 \\ 
%-& \frac{\beta^{(m)}}{N_v^{(m)}} \Big(\sum_i^N c_i^{(m)} (y_i - y_i^{(m)})\Big)^2 \\
%+&\gamma^{(m)} \sum_{i,k} c_i^{(m)}c_k^{(m)} \big((y_k - y_i ) - (y_k^{(m)} - y_i^{(m)})\big)^2 \\
%\end{aligned}
%\end{equation}
%where $\alpha^{(m)} = \alpha_u^{(m)} + \beta^{(m)}$ and therefore $\alpha^{(m)} \geq \beta^{(m)}$.

The inference objective is to find $min_{y} E(y)$. For ease of expression we can re-write \eqref{energy_inference} in the following form:

\begin{equation} 
\label{energy_inference_matrix}
\begin{aligned}
E(y) &= y^{T} A y -2 (y^{T} A^s y^s + y^{T} A^d y^d) 
      \\&+ y^{sT} A^s y^s  + y^{dT} A^d y^d
\end{aligned}
\end{equation}
%The inference objective is to find $min_{\mathbf{y}} E(\mathbf{y})$. For ease of expression we can re-write \eqref{energy_inference} in the following form:
% \begin{equation} 
% \label{energy_inference_matrix}
% \begin{aligned}
% E^{(m)}(y) = \mathbf{y}^T A^{(m)} \mathbf{y}
% &-2 \mathbf{y}^T A^{(m)} \mathbf{y}^{(m)} + \mathbf{y}^{(m)T} A^{(m)} \mathbf{y}^{(m)}
% \end{aligned}
% \end{equation}
where $A=(A^{s}+A^{d})$ is a $N \times N$ symmetric and positive (semi-) definite matrix. $A^{s}$ is a diagonal matrix with entries $A_{ii}^{s}=\alpha c_{i}^{s}, \forall i$ while $A^{d}$ is a dense $N\times N$ symmetric and positive (semi-) definite matrix with entries as follows:
\begin{equation} 
\begin{aligned} [c]
\label{Matrix_A}
A_{ii}^d &= c_i^d \Big{(}\frac{\beta}{N} \textstyle{\sum}_{j,j\neq i}^N \displaystyle c_j^d +\gamma \textstyle{\sum}_{j\in \mathcal{N}(i)} \displaystyle c_j^d\Big{)}  \\
A_{ij}^d &= -c_i^d \Big(\frac{\beta}{N} c_j^d + \gamma c_j^d\Big) \\
A_{ij}^d &= -c_i^d \Big(\frac{\beta}{N} c_j^d\Big) \\
\end{aligned}
\begin{aligned}[c]
\forall i 
\\
\\
\forall i, j \in \mathcal{N}(i) 
\\
 \forall i, j \neq i, j \not\in \mathcal{N}(i)
\end{aligned}
\end{equation}

Differentiating \eqref{energy_inference_matrix} with respect to $y$ and then setting the resulting expression to $0$ we obtain:
\begin{equation} 
\begin{aligned}
\label{derivative_inference}
A y = A^s y^s + A^d y^d
\end{aligned}
\end{equation}

To solve for $y$ in \eqref{derivative_inference} we use the iterative conjugate gradient method. For the algorithm we do not need to explicitly construct matrices $A^s$ and $A^d$ but simply evaluate the gradients $A^sy^s$ and $A^dy^d$ at the start, and $Ay$ at each iteration-step. Note that computing $A^dy^d$ and $Ay$ require $O(N^2)$ operations, however based on the simplified form of equation \eqref{energy_fc_lt}, a linear time expression for gradient computation can be derived.
% \begin{equation} 
% \label{Ay}
% \begin{aligned}
% y_i^{'(m)} &= c_i^{(m)} \Big(\frac{\alpha^{(m)}}{N_v^{(m)}} \textstyle{\sum}_j^N \displaystyle c_j^{(m)}y_i - \frac{\beta^{(m)}}{N_v^{(m)}} \textstyle{\sum}_j^N \displaystyle c_j^{(m)}y_j \\
% &+ \gamma^{(m)} \textstyle{\sum}_{j\in \mathcal{N}(i)} \displaystyle c_j^{(m)}(y_i - y_j)\Big), 
% \end{aligned}
% \end{equation}
% and similarly evaluate $A^{(m)}\mathbf{y}^{(m)}, \forall m$ once at the start. Both of these are linear time operations as the terms $ \sum_j^N c_j^{(m)}$, $\sum_j^N c_j^{(m)}y_j$ and $\sum_j^N c_j^{(m)}y_j^{(m)}$ are common for all $i$ and can be re-used. 
To further accelerate the process we implement the algorithm to run on the GPU, where per pixel operations are parallelized. In practice the solution converges rapidly to within a desired threshold in $n \ll N$ iterations. 

Note that for the system of linear equations to have a unique solution, i.e. non-zero determent for $A$, then $\alpha$ should be non-zero. Intuitively it means that at least one input depth map must contribute to the absolute scale of the fused depth map, else infinite solutions exist where the fused depth map is correct up-to-scale. Also, to prevent a potential condition number of infinity due to diagonal entries in $A$ being equal to $0$, we add a small $\epsilon$ to $c^d$. 

\subsection{Learning to Predict Confidence Weights}
\label{learning}
The goal here is to learn separate CNN models that can model the conditional error distributions of $y^s$ and $y^d$, and predict $c^s$ and $c^d$ respectively, given the respective depth maps and $I$ as input. Since the training setup and network architecture is almost identical for the two cases, for brevity we focus on the training procedure for predicting $c^s$, and mention the differences.

The training loss $L^s$ for predicting $c^s$ is defined as follows:
\begin{equation} 
\label{loss_train}
L^s = \frac{1}{N}\sum_i^N (\hat{c}_i^s - c_i^s)^2
\end{equation}
where $\hat{c}^s$ is the predicted confidence map for $y^s$. Since confidence of a depth value is inversely proportional to its error (and in the more general case scale-invariant error) we define $c_i^s$ as follows:
\begin{equation}
\label{def_confidence}
c_i^s = e^{-\lambda^s|E_i^s|}
\end{equation}
where $\lambda^s>0$ is a tunable parameter controlling the contrast of $c^s$ and $E_i^s$ is the scale-invariant error for a depth value in pixel $i$ defined as:
\begin{equation} 
\label{scale_invariant_depth_error_i}
\begin{aligned}
E_i^s &= (\alpha^s + \beta^s)(y_i^s - y_i) \\ 
&-\frac{\beta^s}{N} \sum_j^N (y_j^s - y_j) \\
&+\gamma^s \sum_{j\in \mathcal{N}(i)} \big((y_j^s - y_i^s) - (y_j - y_i)\big) \\ 
\end{aligned}
\end{equation}
where $y$ here stands for the groundtruth log-depth map. The parameters $(\alpha^s, \beta^s)>0$ can be set based on how scale-invariant we require $E_i^s$ to be ($\alpha^s = 0$ for full-scale-invariance in the case of $y^s$ as it is in a random scale), and $\gamma^s>0$ determines whether the network should emphasize more on learning confidences in local pairwise connectivity.

The inputs to the CNN model are $y^s$ and $I$. The RGB image is first passed through a (9x9 kernel size) convolutional layer with 127 output feature maps. The output feature maps are then concatenated with the input log-depth map and passed through 6 more (5x5 kernel size) convolution layers with each having 128 output feature maps, except for the last layer which regresses the confidence map.  All layers are followed by ReLU activation functions, except the output layer which we leave as linear. At test time the predicted values are clipped between $0$ and $1$ inclusive.

The irregular sparsity structure in $y^s$ poses a difficulty to the learning process, as the network has an additional task of learning which points in the input depth map are valid. This demands higher model capacity. One way to facilitate the learning process is to explicitly model the network to be invariant to the sparsity of the data, for instance by performing masked convolutions at each layer \cite{SparsityInvariantCNN} which require significant computation overhead. 

We believe that a more efficient way to facilitate learning with a small network is to densify the data itself before feeding it into the convolutional layers based on some assumption about the data. Here we opted to perform Delaunay triangulation on the 2D image coordinates corresponding to valid points on the depth map, followed by barycentric-coordinate-based linear interpolation (in inverse depth space) to fill the triangles with log depth values. The latter can be efficiently carried out on the GPU. This densification method is motivated by the fact that most regions in a depth map are typically piecewise-planar. Doing so also enhanced errors in the sparse log depth map that otherwise would have been difficult for the network to pick up. We also found that using the triangulated dense log depth map in place of $y_i^s$ in equation \eqref{scale_invariant_depth_error_i} gave more reliable estimates of $E_i^s$.

In order to get $\hat{c}^s$ from the network output (which is now the predicted confidence map corresponding to the triangulated dense log depth map), we simply perform an element-wise multiplication of the network output with $y^s$'s binary mask. The random variability in map scale of $y^s$ also poses a difficulty to the learning process, and as a solution to this we scale each triangulated dense log depth map so that its mean is equal to the mean of the groundtruth log depths in the entire train set, before passing it into the network. For training, we use the Caffe \cite{Caffe_2014} framework. Training was performed with a batch size of 16, a learning rate of $1e^{-2}$, momentum of 0.9 using SGD as the optimizer on a NVIDIA GTX 1080 Ti GPU.

\section{Experimental Results}
\label{experiments}

This section provides the quantitative and qualitative results of our approach for different experimental settings on NYU Depth v2 \cite{SilbermanECCV12} and KITTI \cite{KITTI} datasets.

%\textcolor{red}{Fixme : other common stuff goes here}. 
\begin{figure}[!t]
\centering
\begin{tabular}{cc}
\includegraphics[width=.45\columnwidth, trim={.1cm 0 1cm 0},clip]{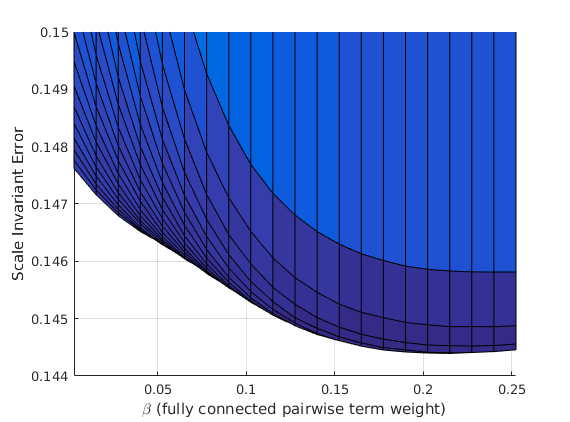}
&\includegraphics[width=.45\columnwidth, trim={.1cm 0 1cm 0},clip]{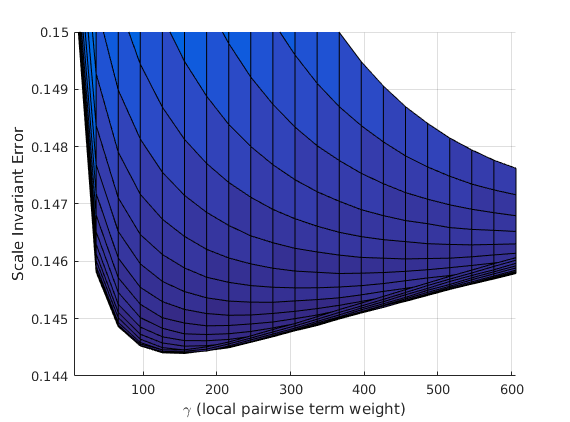}
\end{tabular}
\caption{Plots showing the scale invariant error (on vertical axis) of our just-in-time reconstructions on NYU dataset as we change the hyperparameters (on horizontal axis): $\beta$ (left) and $\gamma$ of our CRF, with $\alpha=10$. Black curve at the bottom in each plot represents best reconstruction error against varied fully connected terms strength and locally connected terms strength respectively.}
\label{fig:hyperparams}
\end{figure}

\begin{table}[!t]
\caption{Quantitative results of inpainting ORB-SLAM maps on NYUv2 dataset.}
\centering
\begin{tabular}{c}

\begin{tabular}{cc}
\midrule
Method & Scale Invariant Error\\ 
\midrule
Sparse Map \cite{ORBSLAM_2015} & 0.492 \\
Eigen et al. \cite{Eigen2014} & 0.159 \\
Ours & \textbf{0.144} \\ 
\hline
%Ablation study\\
%Ours (w. Equal Conf.)  & 0.150  \\ 
%Ours (w. Learned Conf. for Sparse Map) & 0.145 \\ 
%Ours (w. Learned Conf. for Depth Preds.) & 0.149 \\
\end{tabular}
\\

\begin{tabular}{ccc}
\midrule
 \multicolumn{2}{c}{Learned Confidence} & Scale Invariant Error\\ 
 Map & Prediction& \\ 

%\midrule
%Sparse Map  & 0.492 \\
%Eigen \cite{Eigen2014} & 0.159 \\
%Ours & \textbf{0.144} \\ 
\hline
%Ablation study\\
\text{x} & \text{x} & 0.150   \\ 
\text{x} & $\checkmark$ & 0.149 \\
$\checkmark$ & \text{x} & 0.145 \\ 
$\checkmark$ & $\checkmark$ &\textbf{0.144} \\
\end{tabular}

\end{tabular}
\vspace{-1em}
\label{tab:NYU_preds}
\end{table}

\begin{figure*}
\centering
\includegraphics[width=\textwidth]{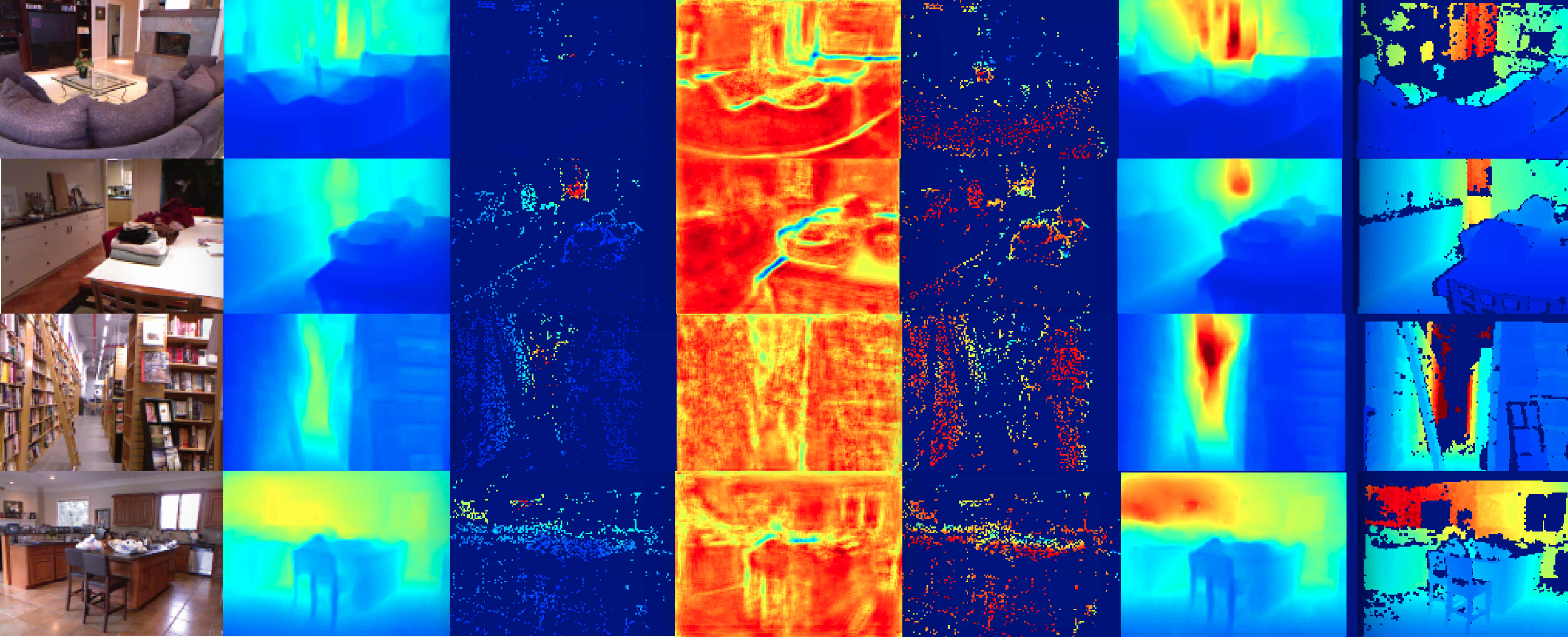}
\caption{ORB-SLAM inpainting results for some of the images in the NYU test set. In column order  $1^{st}$: RGB image, $2^{nd}$: Depth predictions \cite{Eigen2014}, $3^{rd}$: ORB-SLAM sparse depth map \cite{ORBSLAM_2015}, $4^{th}$: Predicted confidences using our method for depth predictions (red implies higher confidence), $5^{th}$: Predicted confidence using our method for the sparse map, $6^{th}$: Our just-in-time reconstruction result, $7^{th}$: Ground truth.}
\label{fig:nyu_results}
\end{figure*}

\subsection{Inpainting Sparse Depth Maps from Multi-View Geometry}

We first demonstrate just-in-time reconstructions of sparse ORB-SLAM maps that are projected onto frames in the subset of the NYU dataset \cite{SilbermanECCV12}. For this experiment, we use the train/test split specified in \cite{Eigen2014}, however our train/test set is a fraction of that of \cite{Eigen2014} limited by the success of ORB-SLAM's \cite{ORBSLAM_2015} tracking on the corresponding scenes in the raw dataset \cite{SilbermanECCV12}. We use the network proposed in \cite{Eigen2014} as the virtual depth sensor for all indoor experiments.

Quantitative results of our just-in-time reconstruction are summarized in Table \ref{tab:NYU_preds} where we report the scale invariant error measure as used in \cite{Ummenhofer2016}. Figure \ref{fig:hyperparams} shows the sensitivity of the just-in-time reconstruction as we vary the strength of the fully connected and the local grid connected terms of the CRF. It is evident that both the terms contribute to the performance in this case.
The first 3 rows of Table \ref{tab:NYU_preds} show that the CRF based inpainting results successfully take advantage of the accumulated geometric evidence from multiple-views contained in the sparse map in order to significantly improve upon the depth prediction errors of the neural network baseline. It is important to note that the ORB-SLAM sparse map scale invariant error measure is computed only on the points visible in the map so does not precisely correspond to that of denser error measures but still indicative that ORB-SLAM maps contain gross outliers.

Figure \ref{fig:nyu_results} shows the qualitative results of our just-in-time reconstruction approach. Dense predictions using \cite{Eigen2014} are often inaccurate at the edges -- a likely artifact of loss in resolution from the single view depth estimation network -- and at regions which are further away from the camera. Our confidence prediction networks most often correctly predicts the likely confidences of the input depth data (even for the sparse ORB-SLAM depth maps which are at arbitrary scale) which help reduce most of the depth errors and gross outliers from making into the fused result.

\begin{figure*}[t]
\centering
\includegraphics[width=\textwidth]{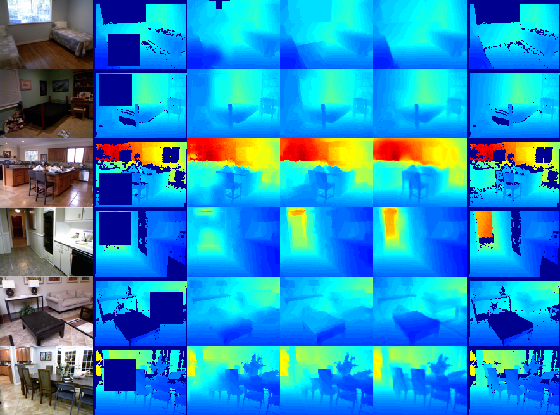}
\caption{Comparison of methods for inpainting Kinect depth maps from the NYU test set with randomly removed data. In column order, $1^{st}$: RGB image, $2^{nd}$: Kinect depth map with randomly removed data, $3^{rd}$: Inpainted depth map using cross-bilateral filtering, $4^{th}$: Inpainted depth map using Colorization, $5^{th}$: Our just-in-time reconstruction result, $6^{th}$: Raw Kinect depth map.}
\label{fig:inpainting}
\end{figure*}

\begin{figure*}
\centering
\includegraphics[width=\textwidth]{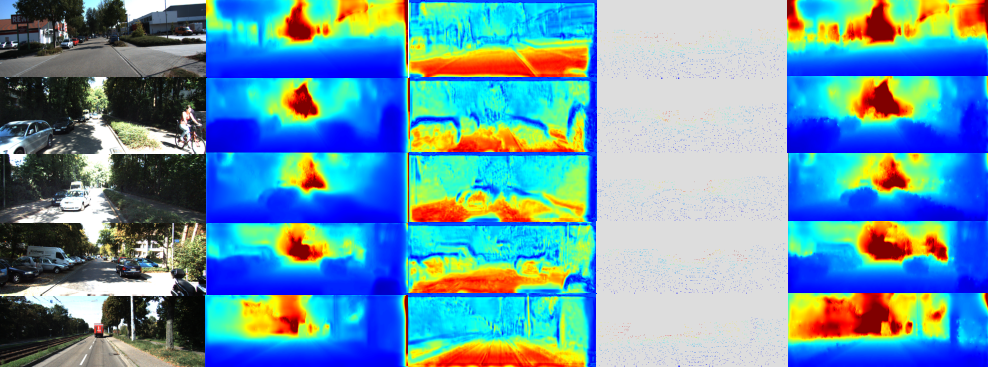}
\caption{Inpainting LIDAR depth maps. For the quantitative evaluation in Table \ref{tab:KITTI_preds} we have randomly removed $2/3^{rd}$ of points from the maps making them even sparser. Notice that in spite of this we are able to densify the maps whilst being consistent with their values and significantly improving over the single view depth predictions. In column order, $1^{st}$: RGB image, $2^{nd}$: Depth predictions \cite{garg2016unsupervised}, $3^{rd}$: Predicted confidences using our method for depth predictions (red denotes higher confidence), $4^{th}$: LIDAR depth map with $2/3^{rd}$ of points randomly removed, $5^{th}$: Our just-in-time reconstruction result.}
\label{fig:LIDAR_inpainting}
\end{figure*}

%\noindent \textbf{Parameter Selection an importance of locally and fully connected layers:}

In Table \ref{tab:NYU_preds}, we provide an ablation study justifying importance of confidence estimation. In the bottom part, we show that incorporating confidences of both the map and the single view depth predictor, is important for obtaining more accurate reconstructions. Dropping confidences estimated in either sparse map or dense depth predictor (or both), as expected, degrades the quality of the just-in-time reconstruction. 
Figure \ref{fig:fusions} shows the same visually. % where the .... (the figure is useless)
Note that interestingly, if we want the final fused result to be closer to metric scale (the scale of the predicted depth map $y^d$) --- while still having similar scale-invariant-error as the \emph{up-to-scale} reconstruction obtained in our current setup --- we can simply use $y^d$ in the unary terms, and $y^s$ in the fully-connected term of our energy formulation (instead of the other way around), and re-tune the hyper-parameters $\alpha$, $\beta$, and $\gamma$.

\begin{table}[!b]
\caption{Kinect (Top)/ LIDAR (Bottom) inpainting results on NYU/KITTI datasets on points removed.}
\centering
\begin{tabular}{cc}
Method &RMSE (m) \\
\midrule
Eigen et al. \cite{Eigen2014} & 0.782 \\
Cross-bilateral Filtering & 0.632 \\
Colorization & 0.497 \\
Ours & \textbf{0.406} \\
\hline
Garg et al. \cite{garg2016unsupervised} & 6.32 \\
%Colorization \cite{}) & \textbf{0.681} \\ 
Ours & \textbf{2.45}
\end{tabular}
\vspace{-1em}
\label{tab:KITTI_preds}
\end{table}

\subsection{Inpainting Sparse Depth Maps from Depth Sensor}

We also evaluate our method for inpainting sparse sensory data captured in indoor and outdoor environments. %In the first set of experiment we use the same subset of NYU as described in the section above\textcolor{red}, and inpaint the Kinect depth-map using \cite{Eigen2014}.
Note that in our experiments for sensory data inpainting, relying purely on our unary and local pairwise terms was sufficient. This is because unlike ORB-SLAM maps, the sensory data used here is fairly accurate with little to no outliers, and thus don't require a fully-connected pairwise neighbourhood of learned depth relationships to rectify the existing sensor data.
Hence, we can fill-in the holes in the sensory depth maps based on the very local depth relationships of the single view depth predictions (only the local pairwise weight $\gamma$ need to be tuned on a validation set for a fixed $\alpha$). On the other hand, if the sensor is noisy and unreliable, and/or if the depth map prediction (or whichever input depth source, whether sparse or dense, that will be used instead for inpainting) is more reliable, then having our fully-connected term would be beneficial, such as in the case for ORB-SLAM inpainting.

We first evaluate our method for inpainting Kinect depth maps on our NYU test subset, again using the network in \cite{Eigen2014} as the virtual depth sensor. 
To quantitatively evaluate the results, while respecting the structured sparsity pattern, we introduce a synthetic version of the NYU test set where a random rectangular region is cropped from the Kinect depth map to be labeled missing and then inpainted using the rest of the visible depth map. In Table \ref{tab:KITTI_preds}, we report the inpainting accuracy only in this missing region to be a reasonable enough quantifier for the accuracy, and compare with depth estimations from \cite{Eigen2014} and commonly used Kinect inpainting methods on the NYU dataset such as cross-bilateral filtering \cite{CrossBilateralFiltering} and Colorization \cite{Colorization} (both taken directly from the NYU toolbox \cite{SilbermanECCV12}) as baselines. Figure \ref{fig:inpainting} shows the qualitative comparison of the results of our CRF based inpainting with that of the baselines. It is clear that the proposed method produce more realistic depth maps as opposed to the mostly piecewise constant and inaccurate depth maps the baselines produce. 

Next we evaluate our method for inpainting sparse LIDAR maps in the KITTI dataset. To facilitate quantitative evaluation we remove $2/3^{rd}$ of the map points (respecting the sparsity structure of the LIDAR data) and evaluate the inpainted results against the removed points. Quantitative results on the test set are shown in Table \ref{tab:KITTI_preds} where our inpainting method improves greatly over the baseline CNN predictions of \cite{garg2016unsupervised}. Some qualitative examples for the same are shown in Figure \ref{fig:LIDAR_inpainting} where we can observe sharper object boundaries in our inpainted result, in comparison to the blurred object boundaries in the predicted depth map due to the loss in resolution that very deep feed forward CNNs suffer from. Most other errors in scene structure in the depth predictions have also been corrected in our fused result.

\subsection{Runtimes}

On our test setup which uses a NVIDIA GTX 980 GPU and Intel i7 4790 CPU, inference time for ORB-SLAM and Kinect depth map inpainting on the NYU dataset is $\approx 30 ms$ at 147x109 image resolution, which is the same resolution as the predicted depth map by \cite{Eigen2014}. Inference time is $\approx 200 ms$ if performed at the full image resolution of 640x480 by upsampling depth predictions. Inference time for LIDAR depth map inpainting on the KITTI dataset is $\approx 100 ms$ at 608x160 image resolution, which is the same resolution as the predicted depth map by \cite{garg2016unsupervised}. Total overhead time for neural network depth and confidence predictions is $\approx 50 ms$ for both datasets. %The $\alpha^{(1)}$,$\alpha^{(2)}$, $\beta^{(1)}$,$\beta^{(2)}$ ,$\gamma^{(1)}$ and $\gamma^{(2)}$ terms were tuned on a validation set were found to be 0.2, 4, 0, 4, 200 and 0 respectively.

\section{Conclusion}
In this work we advocated ``just-in-time reconstruction", a flexible and efficient approach to dense reconstruction that utilizes a sparse map and a higher-level form of scene understanding based on a single live image, to generate a dense reconstruction \emph{on-demand}, that is consistent with the sparse map. This approach is especially useful during large-scale mapping where it's inefficient to store detailed information about the map and all the corresponding images.
% In this work we introduced a simple and efficient means of inpainting sparse maps using a high level form of scene understanding based on an image, a task which we coined as ``just-in-time reconstruction".
We modeled the task as inference over a novel fully connected CRF model, with nodes anchoring the solution to the sparse map and the scale-invariant edges enforcing pairwise depth relationship information based on depth predictions of a deep neural network, given the live RGB image.
The CRF model was also parameterized by point-wise confidences of both the sparse map and dense depth predictions, and these confidences were predicted using CNNs, given the depth maps and the live RGB image as input. This form of probabilistic data fusion allowed the solution to be less sensitive to the presence of erroneous depths, and a simple relaxation of the pairwise confidence weight enabled efficient inference of the solution. We applied our method to perform real-time image-guided inpainting of sparse maps obtained from three different sources: a sparse monocular SLAM framework, Kinect and LIDAR. Our method is very general and applicable for fusing depth information from multiple modalities and thus opens up more useful applications to be explored.

\section{Acknowledgements}
This work was supported by the ARC Laureate Fellowship FL130100102 to IR and the Australian Centre of Excellence for Robotic Vision CE140100016.

\bibliographystyle{IEEEtran}
\bibliography{References}
\small

\end{document}